\documentclass[sn-mathphys]{sn-jnl}% Math and Physical Sciences Reference Style
\jyear{2021}%

\theoremstyle{thmstyleone}%
\theoremstyle{thmstyletwo}%

\theoremstyle{thmstylethree}%

\begin{document}

\title[Article Title]{Depression Symptoms Modelling from Social Media Text: A Semi-supervised Learning Approach}


\author*[1]{\fnm{} \sur{Nawshad Farruque}}\email{nawshad@ualberta.ca}

\author[1]{\fnm{} \sur{Randy Goebel}}\email{rgoebel@ualberta.ca}

\author[2]{\fnm{} \sur{Sudhakar Sivapalan}}\email{sivapala@ualberta.ca}

\author[1]{\fnm{} \sur{Osmar R. Za\"{i}ane}}\email{zaiane@ualberta.ca}

\affil*[1]{\orgdiv{Department of Computing Science}, \orgname{Alberta Machine Intelligence Institute (AMII), Faculty of Science, University of Alberta}, \orgaddress{\city{Edmonton}, \postcode{T6G 2E8}, \state{AB}, \country{Canada}}}

\affil[2]{\orgdiv{Department of Psychiatry}, \orgname{Faculty of Medicine and Dentistry, University of Alberta}, \orgaddress{\city{Edmonton}, \postcode{T6G 2B7}, \state{AB}, \country{Canada}}}

% \affil[3]{\orgdiv{Department}, \orgname{Organization}, \orgaddress{\street{Street}, \city{City}, \postcode{610101}, \state{State}, \country{Country}}}


\abstract{A fundamental component of user-level social media language based clinical depression modelling is depression symptoms detection (DSD). Unfortunately, there does not exist any DSD dataset that reflects both the clinical insights and the distribution of depression symptoms from the samples of self-disclosed depressed population. In our work, we describe a Semi-supervised Learning (SSL) framework which uses an initial supervised learning model that leverages 1) a state-of-the-art large mental health forum text pre-trained language model further fine-tuned on a clinician annotated DSD dataset, 2) a Zero-Shot learning model for DSD, and couples them together to harvest depression symptoms related samples from our large self-curated Depression Tweets Repository (DTR). Our clinician annotated dataset is the largest of its kind. Furthermore, DTR is created from the samples of tweets in self-disclosed depressed users Twitter timeline from two datasets, including one of the largest benchmark datasets for user-level depression detection from Twitter. This further helps preserve the depression symptoms distribution of self-disclosed Twitter users tweets. Subsequently, we iteratively retrain our initial DSD model with the harvested data. We discuss the stopping criteria and limitations of this SSL process, and elaborate the underlying constructs which play a vital role in the overall SSL process. We show that we can produce a final dataset which is the largest of its kind. Furthermore, a DSD and a Depression Post Detection (DPD) model trained on it achieves significantly better accuracy than their initial version.}

%%================================%%
%% Sample for structured abstract %%
%%================================%%

\keywords{Semi-supervised Learning, Zero-Shot Learning, Depression Symptoms Detection, Depression Detection}

\maketitle

\section{Introduction}\label{sec1}
According to Boyd et al. \cite{boyd1982screening}, in developed countries, around 75\% of all psychiatric admissions are young adults with depression. The fourth leading cause of death in young adults is suicide, which is closely related to untreated depression \cite{who_suicide}. Moreover, traditional survey-based depression screening may be in-effective due to the cognitive bias of the patients who may not be truthful in revealing their depression condition. So there is a huge need for an effective, inexpensive and almost real time intervention for depression in this high risk population. Interestingly, among young adults, social media is very popular where they share their day to day activities and the availability of social media services is growing exponentially year by year \cite{o2011impact}. Moreover, according to the research \cite{gowen2012young, naslund2016future, naslund2014naturally}, it has been found that depressed people who are otherwise socially aloof, show increased use of social media platforms to share their daily struggles, connect with others who might have experienced the same and seek help. So, in this research we focus on identifying depression symptoms from a user's social media posts as one of the strategies for early identification of depression. Earlier research confirms that signs of depression can be identified in the language used in social media posts \cite{coppersmith2015clpsych,Choudhury2013RoleSM,de2014mental,Choudhury2013Pred,reece2017forecasting,Rude2004,seabrook2018predicting,yazdavar2017semi,shen2017depression,yadav2020identifying}. Based on this background, linguistic features, such as n-grams, psycholinguistic and sentiment lexicons, word and sentence embeddings extracted from the social media posts can be very useful for detecting depression, especially when compared to other social media related features which are not language specific, such as social network structure of depressed users and their posting behavior. In additoin, the majority of this background research focused on public social media data, i.e., Twitter and Reddit mental health forums for user-level depression detection, because of the relative ease of accessing such datasets (unlike Facebook and other social media which have strict privacy policies). All this background placed emphasis on signs of depression detection, however, they lacked the inclusion of clinical depression modelling; such requires extensive effort in building a depression symptoms detection model. Some of the earlier research has focused on depression symptoms detection, such as: \cite{yazdavar2017semi,yadav2020identifying,mowery2016towards}, they do not attempt to create a clinician-annotated dataset, and later use existing state-of-the-art language models to expand this set. All the previous research does not attempt to curate the possible depression candidate dataset from self-disclosed depressed users timeline. Therefore the main motivation of this work arises from the following:

\begin{enumerate}
     \item \textbf{Clinician-annotated dataset creation from depressed users tweets:} Through leveraging our existing datasets from self disclosed depressed users and learned Depression Post Detection (DPD) model (which is a binary model for detecting signs of depression), we want to curate a clinician-annotated dataset for depression symptoms. This is a more ``in-situ'' approach for harvesting depression symptoms posts compared to crawled tweets for depression symptoms using depression symptoms keywords, 
    %from random users timeline or from those users timeline who have social media profile name carrying depression keywords, 
    as done in most of the earlier literature \cite{mowery2016towards,mowery2017understanding}. We call it in-situ because this approach respects the natural distribution of depression symptoms samples found in the self-disclosed depressed users timeline. Although \cite{yadav2020identifying} collected samples in-situ as well, our clinician-annotated dataset is much bigger and annotation is more rigorous. 
    
    \item \textbf{Gather more data that reflects clinical insight:} Starting from the small dataset found at (1) and a learned DSD model on that, we want to iteratively harvest more data and retrain our model for our depression symptoms modelling or DSD task.
\end{enumerate}

Both our clinician annotated and harvested tweets with signs of depression symptoms is the largest dataset of its kind, to the best of our knowledge.

\section{Methodology}
\label{sec:ssl-method}

To achieve the goals mentioned earlier, we divide our depression symptoms modelling into two parts: (1) \textbf{Clinician annotated dataset Curation:} here we first propose a process to create our annotation candidate dataset from our existing depression tweets from self-disclosed depressed Twitter users. We later annotate this dataset with the help of a clinician amongst others, that helps us achieve our first goal and (2) \textbf{Semi-supervised Learning:} we then describe how we leverage that dataset to learn our first sets of DPD and DSD models and eventually make them robust through iterative data harvesting and retraining or semi-supervised learning \cite{mcclosky2006effective}. 

From our clinician annotated dataset created in step (1), we separate a subset of depression symptoms stratified samples as a test-set. After each step of the SSL process, we report Macro-F1 and Weighted-F1 scores to evaluate the efficacy of that step based on that test-set.

\section{Datasets} 
\label{sec:ssl-datasets}
We create Depression-Candidate-Tweets dataset from the timeline of depressed users in IJCAI-2017 \cite{shen2017depression} who disclosed their depression condition through a self-disclosure statement, such as: "I (am / was / have) been diagnosed with depression" and UOttawa \cite{jamil2017monitoring} datasets where the users were verified by annotators about their ongoing depression episodes. Later, we further filter it with a DPD model (discussed in Section \ref{subsec:ssl-clinician-annot-dataset}) for depression tweets and create the Depression Tweets Repository (DTR) which is used in our SSL process to harvest in-situ tweets for depression symptoms. We also separate a portion of the DTR for clinician annotation for depression symptoms (Figure \ref{fig:dsd-data-curat}).
%We do not use CLPsych-2015 dataset at all for this process, because we want to entirely use that dataset for DTM task later.  In the following sub sections, we describe how different intermediate datasets curation for SSL is done in a step-by-step process.

\subsection{Clinician annotated dataset curation}
\label{subsec:ssl-clinician-annot-dataset}
\begin{figure}%[!h]
\centering
\includegraphics[width=0.8 \textwidth]{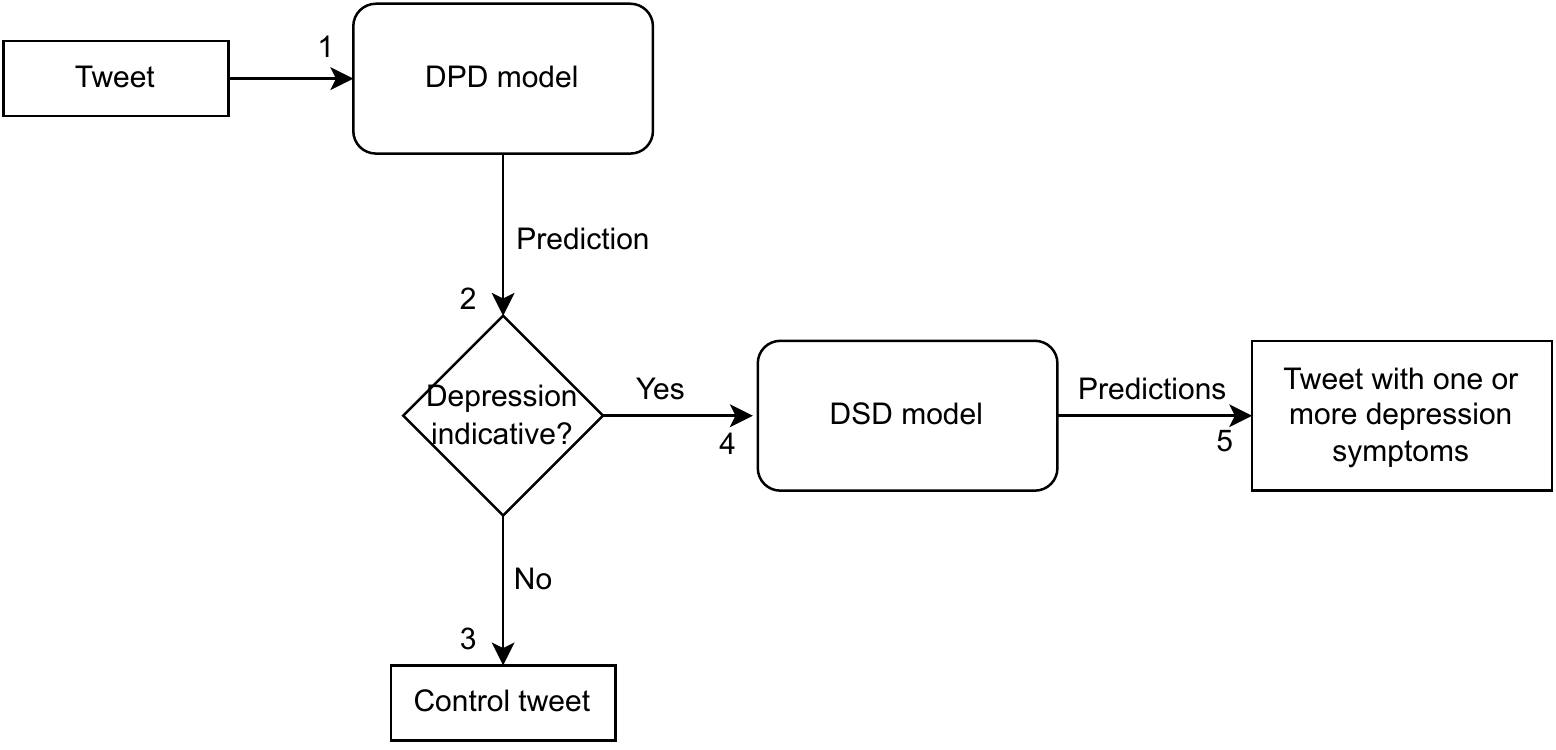}
\caption{\label{fig:dsd-modelling} DSD modelling algorithm}
\end{figure}

\begin{figure}%[!h]
\centering
\includegraphics[width=0.8 \textwidth]{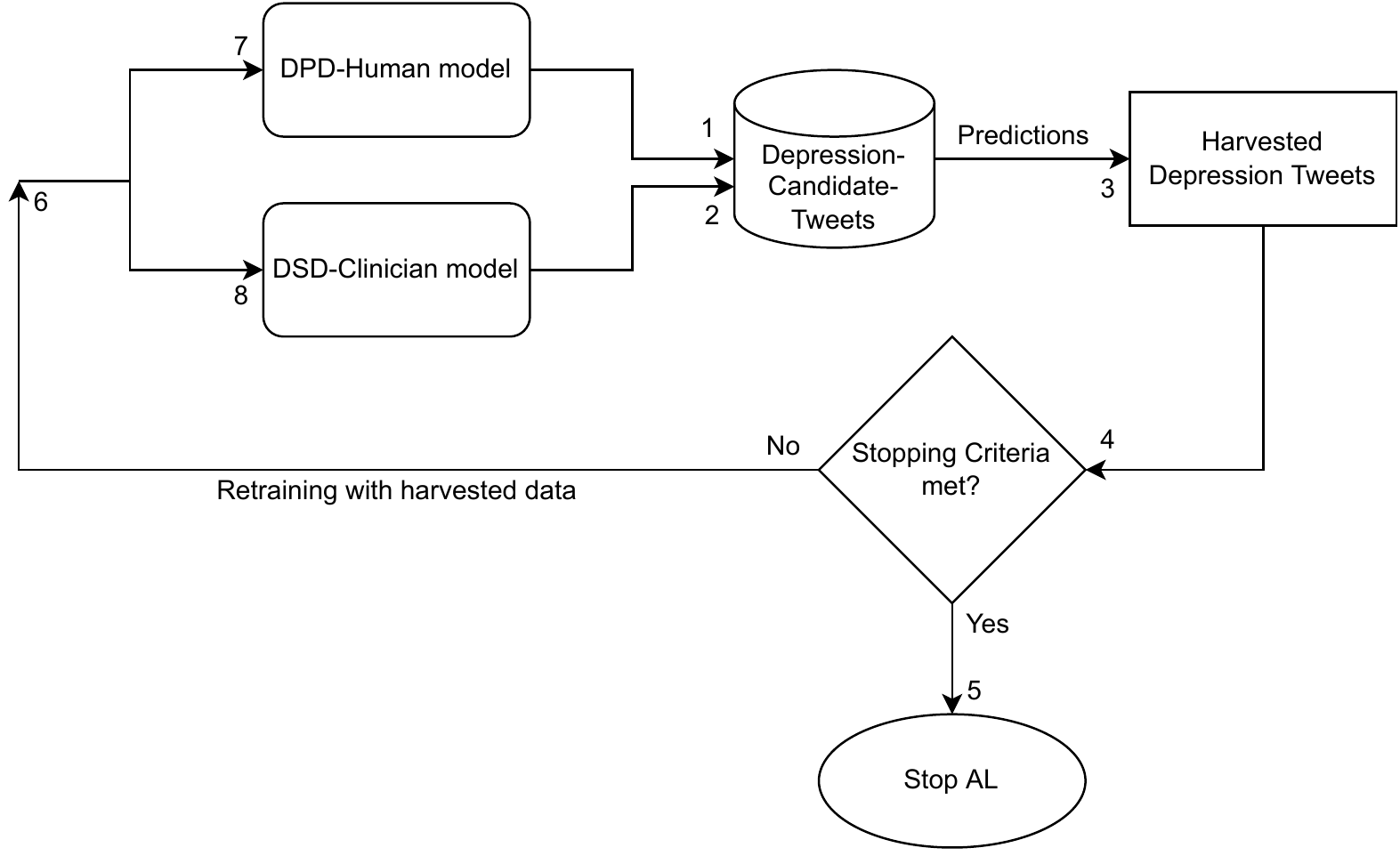}
\caption{\label{fig:brief-ssl} Semi-supervised learning process at a high level.}
\end{figure}

In the overall DSD framework, depicted in Figure \ref{fig:dsd-modelling}, we are ultimately interested in creating a robust DPD and a DSD model which are initially learned on human annotated samples, called ``DPD-Human'' model and ``DSD-Clinician'' model as depicted in Figure \ref{fig:brief-ssl}. The suffixes with these model names, such as ``Human,'' indicates that this model leverages the annotated samples from both non-clinicians and clinicians;  ``Clinician'' indicates that this model leverages the samples for which the clinician's annotation is taken as more important (more explanation is provided later in Section \ref{subsec:ssl-human-annotated-dep-sympts}). At the beginning of this process, we have only a small human annotated dataset for depression symptoms augmented with depression posts from external organizations (i.e. D2S \cite{yadav2020identifying} and DPD-Vioules \cite{vioules2018detection} datasets), no clinician annotated depression symptoms samples, and a large dataset from self-disclosed depressed users (i.e IJCAI-2017 dataset). We take the following steps to create our first set of clinician annotated depression symptoms dataset and DTR which we will use later for our SSL.

\begin{enumerate}
    \item We start the process with the help of a DPD model, which we call DPD Majority Voting model (DPD-MV).
    %to first create candidate depression posts from our depression users social media posts, i.e. Twitter posts. 
    It consists of a group of DPD models \cite{farruque2019augmenting}, where each model leverages pre-trained word embedding (both augmented (ATE) and depression specific(DSE)) and sentence embedding (USE), further trained on a small set of human annotated depression tweets and a Zero-Shot learning model (USE-SE-SSToT). This ZSL model helps determine the semantic similarity between a tweet and all the possible depression symptoms descriptors and returns the top-k corresponding labels. It also provides a score for each label, based on cosine distance. More details are provided in a previous paper \cite{farruque2021explainable}. Subsequently,  the DPD-MV model takes the majority voting of these models for detecting depression tweets. 
    
    \item We then apply DPD-MV on the sets of tweets collected from depressed users timelines (or \textbf{Depression-Candidate-Tweets}, (Figure \ref{fig:dsd-data-curat}) to filter control tweets. The resultant samples, after applying DPD-MV is referred to as Depression Tweet Repository or \textbf{DTR}. We later separate a portion of this dataset, e.g., 1500 depression tweets for human annotation which we call \textbf{DSD-Clinician-Tweets} dataset. Details of the annotation process are described in Section \ref{subsec:ssl-human-annotated-dep-sympts}.
    
     \begin{figure}%[!h]
        \centering
        \includegraphics[width=0.8 \textwidth]{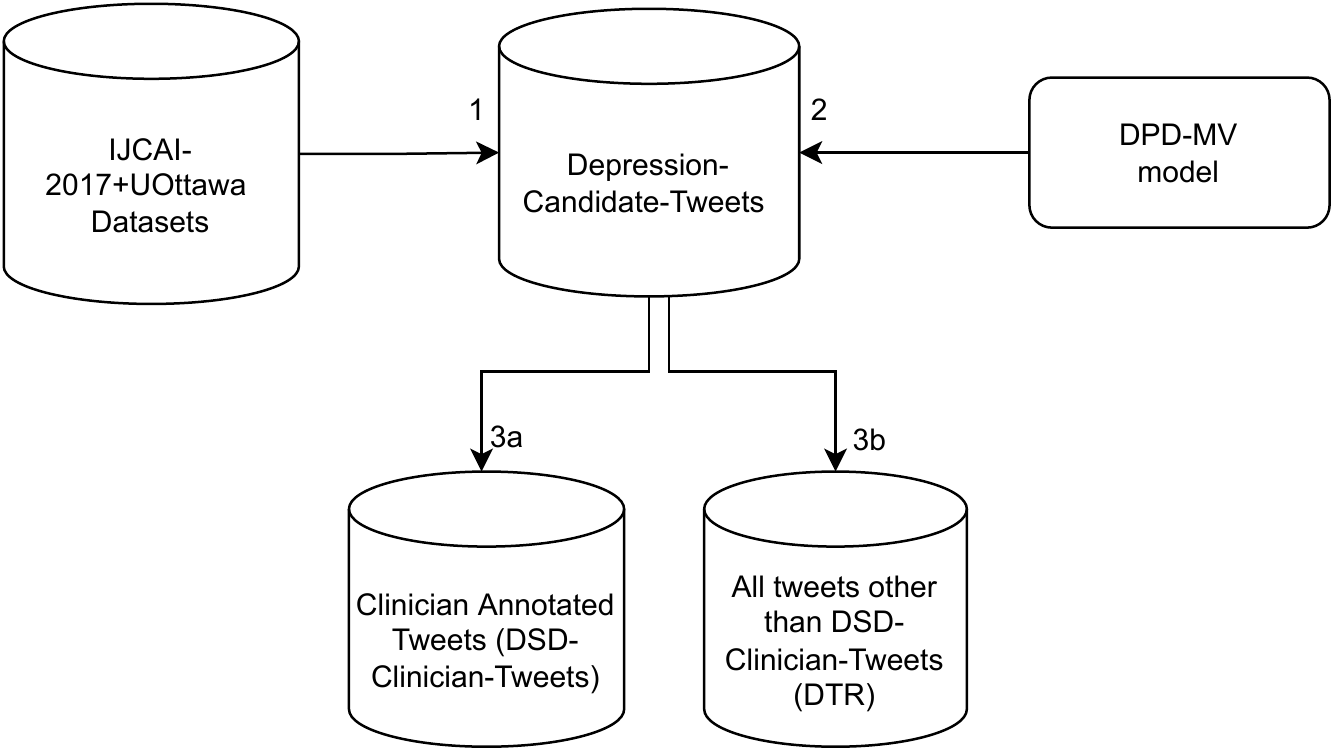}
        \caption{\label{fig:dsd-data-curat} DSD-Clinician-Tweets and DTR curation process}
    \end{figure}
    
    \item We learn our first DSD model using this dataset, then use this model to harvest more samples from DTR.
    %and external organization curated data for depression symptoms. 
    An outline of the DTR and DSD-Clinician-Tweets curation process is provided in Figure \ref{fig:dsd-data-curat}. We describe the details of this process in the Semi-supervised Learning section, but describe each of its building blocks in the next sections. In Table \ref{tab:datasets_for_annot_curat} we provide relevant datasets description.
    
    \begin{table}%[!h]
    \centering
        \begin{tabular}{|c|c|p{6cm}|}
        \hline
             Dataset &Sample size  &Comment \\
             \hline
             Depression-Candidate-Tweets &42,691 & Depressed users' tweets\\
             DTR &6,077 &Depression tweets repository\\
             DSD-Clinician-Tweets &1,500 &Clinician annotated tweets\\
             \hline
        \end{tabular}
        \caption{\label{tab:datasets_for_annot_curat}Datasets}
    \end{table}
    
\end{enumerate}

\subsection{Annotation Task Description}\label{subsec:ssl-annot-task-desc}
Our annotation task consists of labelling a tweet for either 1) one or more of 10 symptoms of depression, 2) No Evidence of Depression(NoED), 3) Evidence of Depression(ED) or 4) Gibberish. We have 10 labels instead of the traditional nine depression symptoms labels because we separate the symptom ``Agitation / Retardation'' into two categories so that our model can separately learn and distinguish these labels, unlike previous research \cite{yadav2020identifying}. NoED indicates the absence of any depression symptoms expressed in a tweet. ED indicates multiple symptoms of depression expressed in a tweet in a way so that it's hard to specifically pinpoint these combined depression symptoms in that tweet. Gibberish is a tweet less than three words long and, due to the result of crawling or data pre-processing, the tweet is not complete and it's hard to infer any meaningful context.

\subsection{Annotation Guideline Creation}\label{subsec:ssl-annot-guideline}
To create the annotation guideline for the task, we analyze the textual descriptions of depression symptoms from most of the major depression rating scales, such as, PHQ-9, CES-D, BDI, MADRS and HAM-D \cite{national2010classification}. We also use DSM-5 as our base reference for symptoms description. Based on these descriptions of the symptoms from these resources and several meetings with our clinicians, we consolidate some of the most confusing samples of tweets from DTR and map them to one or more of those depression symptoms. We then create an annotation guideline with clear description of the clinical symptoms of depression that an annotator should look for in the tweets followed by relevant tweets examples for them including the confusing ones previously noted. We then separate a portion of 1500 samples from our DTR and provide it to the annotators along with our annotation guideline. During the annotation we randomly assign a set of tweets multiple times to calculate test-retest reliability scores. We find annotators annotate the tweets consistently with the same annotation with 83\% reliability based on the test-retest reliability score. Our detailed guideline description is provided in Appendix \ref{appen:annots-guideline}.

\subsection{Depression Symptoms Annotation Process}
\label{subsec:ssl-human-annotated-dep-sympts}
We provide a portion of 1500 tweets from DTR for depression symptoms annotation by four annotators. 
%All the four annotators are thirty plus aged and has a graduate degree. Three of the annotators are native English speakers and one speaks English as a second language. 
Among these annotators two have clinical understanding of depression: one is a practising clinician and the other one has a PhD in Psychiatry.

In our annotation process, we emphasize the  annotating of a tweet based on the clinical understanding of depression which is laid out in our annotation guideline. We take majority voting to assign a label for the tweet. In absence of majority, we assign a label based on the clinician's judgment, if present, otherwise, we do not assign a label to that tweet. We call this scheme \textbf{Majority Voting with Clinician Preference (MVCP)}. Table \ref{tab:pairwise-kappa-scores} reports the average kappa scores for each labels and Annotator-Annotator, Annotator-MVCP and All pairs (i.e. avg. on both of the previous schemes).

%We provide pair wise kappa agreement score for all pairs of annotators and for each of our labels. We also provide each of the annotators agreement with the inferred labels using MVCP. See Table \ref{tab:pairwise-kappa-scores}.

We observe fair to moderate kappa agreement score (0.38 - 0.53) among our annotators for all the labels. We also find, ``Suicidal thoughts'' and ``Change in Sleep Patterns'' are the labels for which inter-annotator agreement is the highest and agreement between each annotator and MVCP is substantial for the same. Among the annotators the order of the labels based on descending order of agreement score is as follows: Suicidal thoughts, Change in Sleep Patterns, Feelings of Worthlessness, Indecisiveness, Anhedonia, Retardation, Weight change, NoED, Fatigue, Low mood, Gibberish, Agitation and ED. However, with MVCP, we find moderate to substantial agreement (0.56 - 0.66).

\begin{sidewaystable}%[!h]
    %\footnotesize
    \centering
    \begin{tabular}{|l|l|l|l|}
    \hline
        \textbf{Depression-Symptom-Labels} & \textbf{Average(Annots.)} & \textbf{Average(Annots.- MVCP)} & \textbf{Average(All)} \\ \hline
        Suicidal thoughts                   &$0.5319(\pm{0.1045})$ &$0.6296(\pm{0.1227})$ &$0.5710(\pm{0.1167})$ \\ \hline
        Change in Sleep Pattern             &$0.5171(\pm{0.0770})$ &$0.6162(\pm{0.1034})$ &$0.5568(\pm{0.0973})$ \\ \hline
        Feelings of Worthlessness           &$0.4517(\pm{0.1978})$ &$0.6589(\pm{0.2347})$ &$0.5346(\pm{0.2271})$ \\ \hline
        Indecisiveness                      &$0.4475(\pm{0.2164})$ &$0.6378(\pm{0.2479})$ &$0.5236(\pm{0.2370})$ \\ \hline
        Anhedonia                           &$0.4434(\pm{0.2383})$ &$0.6037(\pm{0.0915})$ &$0.5076(\pm{0.2030})$ \\ \hline
        Retardation                         &$0.4382(\pm{0.3030})$ &$0.5672(\pm{0.2446})$ &$0.4898(\pm{0.2746})$ \\ \hline
        Weight Change                       &$0.4358(\pm{0.1589})$ &$0.6155(\pm{0.2149})$ &$0.5077(\pm{0.1951})$ \\ \hline
        NoED  &$0.4321(\pm{0.2119})$ &$0.5946(\pm{0.2631})$ &$0.4971(\pm{0.2346})$ \\ \hline
        Fatigue                             &$0.4297(\pm{0.1136})$ &$0.5975(\pm{0.2375})$ &$0.4968(\pm{0.1830})$ \\ \hline
        Low Mood                            &$0.4251(\pm{0.3041})$ &$0.6454(\pm{0.3730})$ &$0.5132(\pm{0.3327})$ \\ \hline
        Gibberish                           &$0.4172(\pm{0.2606})$ &$0.6626(\pm{0.3272})$ &$0.5154(\pm{0.2991})$ \\ \hline
        Agitation                           &$0.4008(\pm{0.2066})$ &$0.6505(\pm{0.2571})$ &$0.5007(\pm{0.2498})$ \\ \hline
        ED     &$0.3877(\pm{0.0878})$ &$0.5765(\pm{0.2742})$ &$0.4632(\pm{0.1971})$ \\ \hline
    \end{tabular}
    \caption{\label{tab:pairwise-kappa-scores} Pairwise kappa scores among annotators and MVCP for all the labels}
\end{sidewaystable}

\subsection{Distribution Analysis of the Depression Symptoms Data}\label{subsec:ssl-dsd-sympts-distr}

\begin{figure}%[!h]
    \centering
    \includegraphics[width=1 \textwidth]{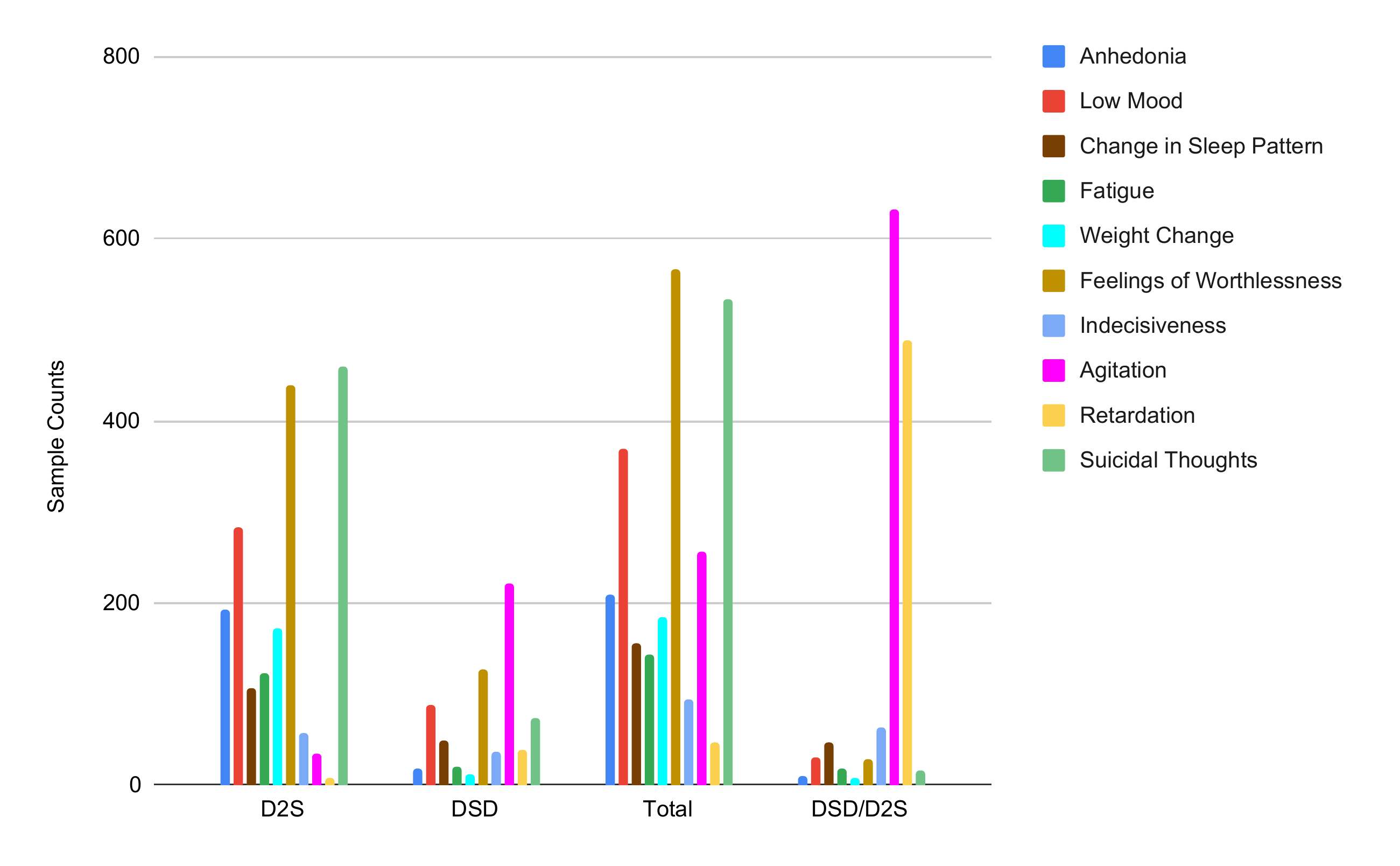}
    \caption{\label{fig:sympt-distr} Sample distribution and ratio analysis across D2S and DSD datasets}
\end{figure}

In this section we provide symptoms distribution analysis for our D2S and DSD-Clinician-Tweets datasets. DSD-Clinician-Tweets dataset contains 1500 tweets. We then create a clean subset of this dataset which holds clinicians annotations and only tweets with depression symptoms,  which we call DSD-Clinician-Tweets-Original (further detail is in Section \ref{subsubsec:ssl-first-dsd-clinician}). For D2S, we have 1584 tweets with different depression symptoms labels. In Figure \ref{fig:sympt-distr}, the top 3 most populated labels for DSD dataset are Agitation, Feeling of Worthlessness and Low Mood. However, for D2S dataset Suicidal Thought is the most populated label followed by Feeling of Worthlessness and Low Mood, just like DSD. We use D2S dataset because D2S crawled tweets from self-reported depressed users timeline. Although they did not confirm whether these users have also disclosed their depression diagnosis, they mention that they analyze their profile to ensure that these users are going through depression. Since their annotation process is not as rigorous as ours, i.e., they did not develop an annotation guideline as described in the earlier section and their depressed users dataset may not contain all self-disclosed depressed users, we had to further filter those tweets before we could use them. So we use DSD-Clinician-Original-Tweets for training our very first model in SSL process, later use that to re-label D2S samples.
%This observation signifies the fact that naturally samples with Suicidal Thoughts label are not in abundant in the depression tweets by the self-disclosed depression users. Artificial spike of Suicidal Thoughts label is possible due to lexicon or keyword based data crawling where the crawler fetches these samples from possibly fake profiles. Also, it is possible to build models using such datasets which is particularly good for that label although in practical situation samples for that label is not that abundant. 
In a later section we report the distribution on harvested data and another approach for increasing sample size for least populated labels. 

\section{Data Preprocessing} \label{sec:ssl-data-preproc}
We perform the following preprocessing steps for all our Twitter datasets, we use NLTK \footnote{https://www.nltk.org/book/ch06.html} for tokenizing our tweets and also Ekphrasis \footnote{https://github.com/cbaziotis/ekphrasis} for normalizing tweets.

\begin{enumerate}
    \item Lowercase each words.
    % \item Remove re-tweets and replies.
    \item Remove one character words and digits.
    %\item Remove tweets which are less than three words long.
    \item Re-contract contracted words in a tweet. For example, ``I've'' is made ``I have''.
    \item Elongated words are converted to their original form. For example, ``Looong'' is turned to ``Long''.
    \item Remove tweets with self-disclosure, i.e. any tweet containing the word ``diagnosed'' or ``diagnosis'' is removed.
    \item Remove all punctuations except period, comma, question mark and exclamation. 
    %Punctuation have been found useful to represent a text based on sentence embedding.
    \item Remove URLs.
    \item Remove non-ASCII characters from words.
    \item Remove hashtags
    \item Remove emojis.
\end{enumerate} 

\section{Experimental Setup and Evaluation}
\label{sec:ssl-exp-setup}
Our experimental setup consists of iterative data harvesting and re-training of a DSD model, followed by observing its accuracy increase over each iteration coupled with incremental initial dataset size increase. We report the results separately for each of the steps of SSL in next sections. DSD is a multi-class multi-label problem. We report accuracy measure in Macro-F1 and Weighted-F1. Macro-F1 is an average F1 score for all the labels, where weighted F1-score is a measure which assigns more weight to the labels for which we have most samples. 

\begin{figure}%[!ht]
\centering
\includegraphics[width=1.0 \textwidth]{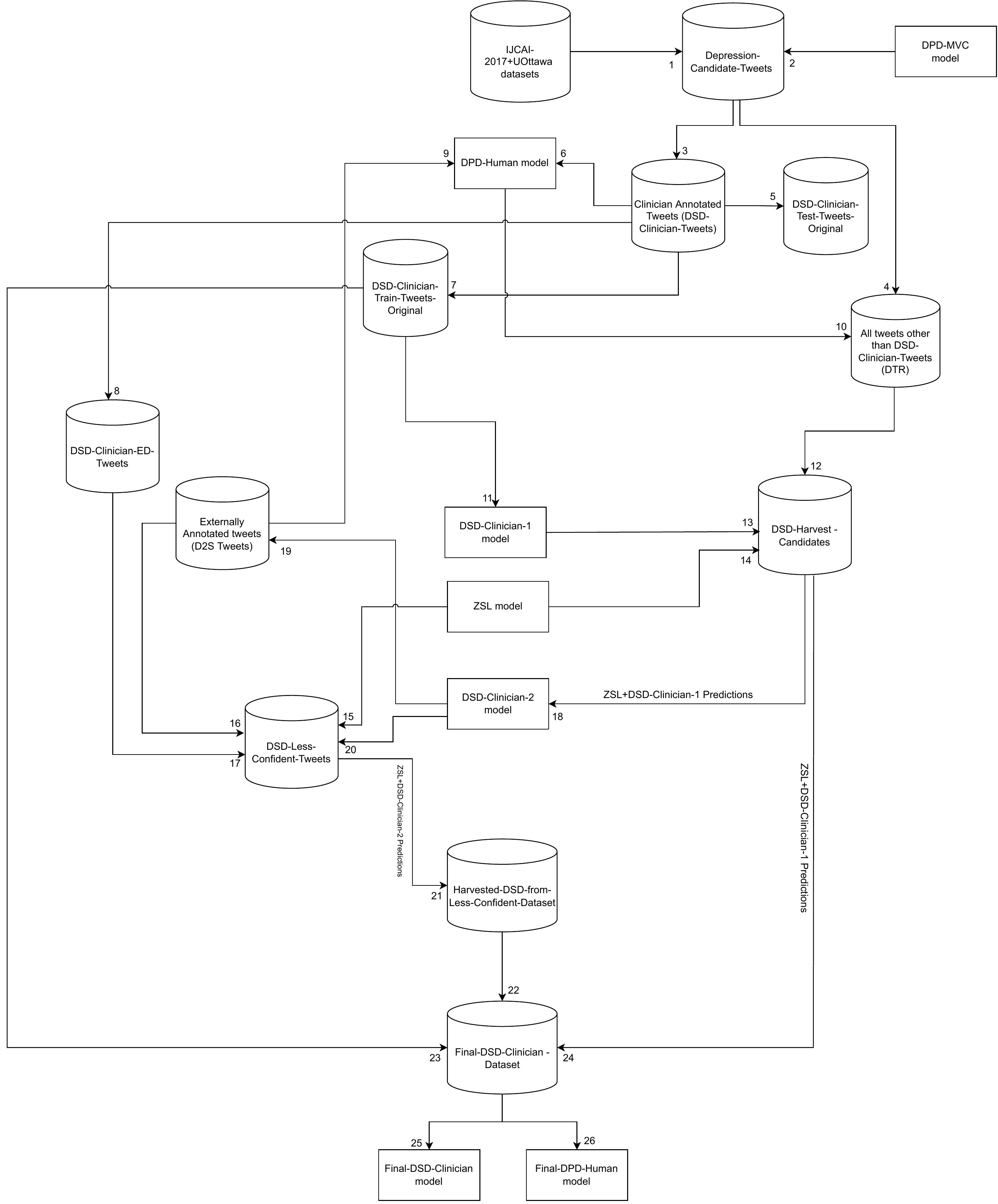}
\caption{\label{fig:detailed-ssl}Detailed semi-supervised learning framework. Here, we show the interaction among our datasets and models. Datasets are shown as cylinders, models are shown as rectangles. An arrow from a dataset to another dataset represents data subset creation; an arrow to another model means the provision of training data for that model; and an arrow from a model to a dataset means use of that model to harvest samples from the dataset. All the arrow heads are marked, so that these can be easily referred while describing a particular scenario in the SSL framework.}
\end{figure}

Our Semi-supervised Learning (SSL) strategy uses the DPD and DSD models and the datasets as described in earlier sections to iteratively harvest more relevant samples and learn robust models (Figure \ref{fig:detailed-ssl}). 

%In that figure, we show the interaction among our datasets and models. Datasets are shown as cylinders, models are shown as rectangles. Arrow from a dataset to another dataset means data subset creation, an arrow to another model means train data for the model and an arrow from a model to a dataset means use of that model to harvest samples from the dataset. 

%In exceptional cases predictions from a particular dataset of a model that is used to create another dataset is also depicted by an arrow from a cylinder to another cylinder, however, in this case this function is described over the arrow, see arrow 21 and 24 in the figure. 
% All the arrow heads are labelled to enable convenient reference while describing a particular scenario in the SSL framework.

\subsection{Semi-supervised Learning (SSL) framework} \label{subsec:ssl-framework}
In our SSL framework, we iteratively perform data harvesting and retraining of our DSD model, which is a multi-label text classifier utilizing pre-trained Mental-BERT \footnote{\url{https://huggingface.co/mental/mental-bert-base-uncased}}, technical details of this model (i.e., the training hyper-parameters) is provided in the Appendix \ref{appen:train-params}. We find Mental-BERT based DSD performs significantly better in terms of Macro-F1 and Weighted-F1 scores compared to base BERT only models in the DSD task (Tables \ref{tab:dsd_clinician_1_bert_model_accuracy} and \ref{tab:dsd_clinician_1_model_accuracy}). In this section, we provide our step by step SSL process description, datasets utilized at each step and the resulting models and/or datasets. 
%We also take few decisions on how to proceed at the next stage based on some observation in the current stage. 
All our steps are depicted in points 11-25 in Figure \ref{fig:detailed-ssl} and described further below.
%the points (11-25) in Figure \ref{fig:detailed-ssl} and described below. 

\subsubsection{Step 1: Creating first DSD model} \label{subsubsec:ssl-first-dsd-clinician}
 In this step, we focus on the creation of a training dataset and a test dataset selected from our clinician annotated samples. This dataset consists of tweets carrying at-least one of the 10 depression symptoms. We use this training dataset to create our first DSD model, called \textbf{DSD-Clinician-1}. To do so, we follow the steps stated below.

\begin{enumerate}
    \item We first remove all the tweets with labels ``Gibberish,'' ``Evidence of Depression'' (ED) and ``No Evidence of Depression'' (NoED) from a subset of  DSD-Clinician-Tweets after applying MVCP. We call this dataset \textbf{DSD-Clinician-Tweets-Original}. Details of ED, NoED and Gibberish are provided in Table \ref{tab:init_datasets}.
    
    \item  We save the tweets labelled as ``Evidence of Depression,'' which we call \textbf{DSD-Clinician-ED-Tweets}, (Arrow 8 in Figure \ref{fig:detailed-ssl}). We later use those to harvest depression symptoms related tweets. 
    
    \item Next, we separate 70\% of the tweets from DSD-Clinician-Tweets-Original dataset and create \textbf{DSD-Clinician-Tweets-Original-Train} dataset for training our first version of DSD model, called \textbf{DSD-Clinician-1} and the rest 30\% of the tweets are used as an SSL evaluation set, also called, \textbf{DSD-Clinician-Tweets-Original-Test}, (Arrows 5 and 7 in the Figure \ref{fig:detailed-ssl}). We will use this evaluation set all through our SSL process to measure the performance of SSL, i.e., whether it helps increase accuracy for DSD task or not. We report the datasets created in this step in Table \ref{tab:init_datasets}, models in Table \ref{tab:init_models} and accuracy scores for each labels and their average in Table \ref{tab:dsd_clinician_1_model_accuracy}. We also report accuracy for the DPD-Human model in this step in Table \ref{tab:dpd_human_model_accuracy}.
    
    \begin{table}%[!h]
    \centering
        \begin{tabular}{|c|c|p{3cm}|}
        \hline
             Dataset &Sample size  &Comment \\
             \hline
             DSD-Clinician-Tweets-Original &539 & Tweets with depression symptoms only\\
             DSD-Clinician-Tweets-Original-Train &377 &Initial train dataset \\
             DSD-Clinician-Tweets-Original-Test &162 &Overall test dataset \\
             DSD-Clinician-ED-Tweets &135 &Depression tweets \\
             DSD-Clinician-NoED-Tweets &785 &Control tweets \\
             DSD-Clinician-Gibberish-Tweets &41 &Gibberish tweets \\
             \hline
        \end{tabular}
        \caption{\label{tab:init_datasets}Datasets in step 1}
    \end{table}

    \begin{table}%[!h]
        \centering
        \begin{tabular}{|c|p{4cm}|p{2cm}|p{3cm}|}
        \hline
             Model &Train dataset &Sample size   &Comment\\
             \hline
             DSD-Clinician-1 &DSD-Clinician-Tweets-Original-Train &377 &DSD-Clinician model at SSL iteration 1\\
             DPD-Human & (DSD-Clinician-Tweets + D2S \text{--} (DSD-Gibberish-Tweets + DSD-NoED-Tweets + Tweets with self-disclosure)) + equal number of NoED tweets from DTR &$(1500+1584-(785+41+34)) + 2224 = 4448$ &DPD-Human model at SSL iteration 1\\
             \hline
        \end{tabular}
        \caption{\label{tab:init_models}Model details in step 1}
    \end{table}

    \begin{table}%[!h]
        % \small
        \centering
        \begin{tabular}{|p{5cm}|c|c|c|c|}
        \hline
            Comment                  &Precision  &Recall &F1-score   &Support\\
            \hline
            Anhedonia                 &0.00      &0.00      &0.00         &5\\
            Low mood                  &0.00      &0.00      &0.00        &26\\
            Change in sleep pattern   &1.00      &0.07      &0.12        &15\\
            Fatigue                   &0.00      &0.00      &0.00         &6\\
            Weight change             &0.00      &0.00      &0.00         &4\\
            Feelings of worthlessness &0.55      &0.16      &0.24        &38\\
            Indecisiveness            &0.00      &0.00      &0.00        &11\\
            Agitation                 &0.55      &0.73      &0.62        &66\\
            Retardation               &0.00      &0.00      &0.00        &12\\
            Suicidal thoughts         &1.00      &0.14      &0.24        &22\\
            \hline
                        %Micro avg     &0.62      &0.51      &0.56       &205\\
                        Macro avg     &0.31      &0.11      &0.12       &205\\
                        Weighted avg  &0.46      &0.28      &0.28       &205\\
                        %Samples avg   &0.60      &0.57      &0.57       &205\\
            \hline
        \end{tabular}
        \caption{\label{tab:dsd_clinician_1_bert_model_accuracy}DSD-Clinician-1 (BERT) model accuracy}
    \end{table}

    \begin{table}%[!h]
        % \small
        \centering
        \begin{tabular}{|p{5cm}|c|c|c|c|}
        \hline
            Comment                  &Precision  &Recall &F1-score   &Support\\
            \hline
            Anhedonia                 &0.00      &0.00      &0.00         &5\\
            Low mood                  &0.61      &0.42      &0.50        &26\\
            Change in sleep pattern   &0.76      &0.87      &0.81        &15\\
            Fatigue                   &0.00      &0.00      &0.00         &6\\
            Weight change             &0.00      &0.00      &0.00         &4\\
            Feelings of worthlessness &0.49      &0.53      &0.51        &38\\
            Indecisiveness            &0.00      &0.00      &0.00        &11\\
            Agitation                 &0.63      &0.77      &0.69        &66\\
            Retardation               &0.00      &0.00      &0.00        &12\\
            Suicidal thoughts         &0.91      &0.45      &0.61        &22\\
            \hline
                        %Micro avg     &0.62      &0.51      &0.56       &205\\
                        Macro avg     &0.34      &0.30      &0.31       &205\\
                        Weighted avg  &0.52      &0.51      &0.51       &205\\
                        %Samples avg   &0.60      &0.57      &0.57       &205\\
            \hline
        \end{tabular}
        \caption{\label{tab:dsd_clinician_1_model_accuracy}DSD-Clinician-1 model accuracy in step 1}
    \end{table}
    
     \begin{table}%[!h]
        % \small
        \centering
        \begin{tabular}{|c|c|c|c|}
        \hline
                Precision  &Recall &F1-score   &Support\\
        \hline
                0.84       &0.90   &0.87       &227 \\
        \hline
        \end{tabular}
        \caption{\label{tab:dpd_human_model_accuracy}DPD-Human model accuracy in step 1}
    \end{table}
    
\end{enumerate}

\subsubsection{Step 2: Harvesting tweets using  DSD-Clinician-1} \label{subsubsec:ssl-harvest-dsd-clinician}
In this step, we use DSD-Clinician-1 model created in the previous step to harvest tweets which carry signs of depression symptoms from a set of tweets filtered for carrying signs of depression only by \textbf{DPD-Human} model from DTR, we call this dataset \textbf{DSD-Harvest-Candidate-Tweets} (Arrows 10 and 12 in Figure \ref{fig:detailed-ssl}). Our DPD-Human model is trained on all available human annotated datasets, i.e., DSD-Clinician-Tweets-Original and D2S tweets and equal number of control tweets from DTR (Arrows 6 and 9 in Figure \ref{fig:detailed-ssl}) and more dataset details in Table \ref{tab:init_models}. We use this model to leverage human insights to further filter DTR. In this step, we create two more datasets from DSD-Harvest-Candidate-Tweets, (1) \textbf{Harvested-DSD-Tweets:} This dataset contains the tweet samples for which the model is confident, i.e., it detects one of the 10 depression symptoms and (2) \textbf{Harvested-DSD-Tweets-Less-Confident:} This dataset contains the tweet samples for which the model has no confident predictions or it does not predict any depression symptoms for harvested dataset statistics (Table \ref{tab:harvested_datasets_1}).

\begin{table}%[!h]
\centering
    \begin{tabular}{|p{5cm}|p{2cm}|p{2.5cm}|}
    \hline
         Dataset &Sample size  &Comment\\
        \hline
        DSD-Harvest-Candidate-Tweets &3145 &Harvestable tweets for DSD\\
        Harvested-DSD-Tweets &2491 &First harvested dataset\\
        Harvested-DSD-Tweets-Less-Confident &654 &First harvested less confident dataset\\
        \hline
    \end{tabular}
    \caption{\label{tab:harvested_datasets_1}Datasets in step 2}
\end{table}

\begin{table}%[!h]
    % \small
    \centering
    \begin{tabular}{|p{5cm}|c|c|c|c|}
    \hline
        Comment                  &Precision  &Recall &F1-score   &Support\\
        \hline
        Anhedonia                 &0.00      &0.00      &0.00         &5\\
        Low mood                  &0.71      &0.46      &0.56        &26\\
        Change in sleep pattern   &0.70      &0.93      &0.80        &15\\
        Fatigue                   &0.00      &0.00      &0.00         &6\\
        Weight change             &0.00      &0.00      &0.00         &4\\
        Feelings of worthlessness &0.44      &0.63      &0.52        &38\\
        Indecisiveness            &0.00      &0.00      &0.00        &11\\
        Agitation                 &0.62      &0.77      &0.69        &66\\
        Retardation               &0.00      &0.00      &0.00        &12\\
        Suicidal thoughts         &0.80      &0.55      &0.65        &22\\
        \hline
                    %Micro avg     &0.62      &0.51      &0.56       &205\\
                    Macro avg     &0.33      &0.33      &0.32       &205\\
                    Weighted avg  &0.51      &0.55      &0.52       &205\\
                    %Samples avg   &0.60      &0.57      &0.57       &205\\
        \hline
    \end{tabular}
    \caption{\label{tab:harvested_dsd_model_accuracy}DSD-Clinician-1 model accuracy in step 2}
\end{table}

\subsubsection{Step 3: Harvesting tweets using best ZSL Model} \label{subsubsec:ssl-harvest-zsl}
In this step, we use a ZSL model (USE-SE-SSToT) described in \cite{farruque2021explainable} to harvest tweets carrying signs of depression symptoms from the DSD-Harvest-Candidate-Tweets. We choose this model because it has reasonable accuracy in the DSD task and it is fast. We also set a threshold while finding semantic similarity between the tweet and the label descriptor to be more on a conservative side so that we reduce the number of false positive tweets. We find that a threshold $< 1$ is a reasonable choice because cosine-distance $< 1$ indicates higher semantic similarity. In this step, we create two datasets: (1) \textbf{Only-ZSL-Pred-on-Harvested-DSD-Tweets (step: 3a):} This dataset is only ZSL predictions on DSD-Harvest-Candidate-Tweets. (2) \textbf{ZSL-and-Harvested-DSD-Tweets (step: 3b):} This dataset is a combination of ZSL predictions and DSD-Clinician-1 predictions on DSD-Harvest-Candidate-Tweets. 
%create Only-ZSL-Pred-on-Harvested-DSD-Tweets dataset 
We follow steps: 3a and 3b to compare whether datasets produced through these steps help in accuracy gain after using them to retrain DSD-Clinician-1. 

% Compared to step 1 (Table \ref{tab:dsd_clinician_1_model_accuracy}), we achieve 4\% gain in Macro-F1 and 5\% gain in Weighted-F1 using the combined dataset in step: 3b (Table \ref{tab:zsl_and_harvested_dsd_model_accuracy}), compared to 1\% in both the measures using DSD-Harvested-Tweets only in step: 2 and with ZSL only in step: 3a (Table \ref{tab:only_zsl_on_harvested_dsd_model_accuracy}). In step 3a with ZSL only harvesting, we actually lose 3\% in Macro-F1 and 15\% in Weighted-F1. We also provide our produced datasets description in Table \ref{tab:harvested_datasets_2}.  

Compared to step 1 (Table \ref{tab:dsd_clinician_1_model_accuracy}), we achieve 4\% gain in Macro-F1 and 5\% gain in Weighted-F1 using the combined dataset in step: 3b (Table \ref{tab:zsl_and_harvested_dsd_model_accuracy}). We achieve 1\% gain in both the measures using DSD-Harvested-Tweets only in step: 2. With ZSL only in step: 3a (Table \ref{tab:only_zsl_on_harvested_dsd_model_accuracy}), we actually lose 3\% in Macro-F1 and 15\% in Weighted-F1. We also provide our produced datasets description in Table \ref{tab:harvested_datasets_2}. 

%we actually lose 3\% in Macro-F1 and 15\% in Weighted-F1. We also provide our produced datasets description in Table \ref{tab:harvested_datasets_2}. 

\begin{table}%[!h]
\centering
    \begin{tabular}{|p{3cm}|p{2cm}|p{5cm}|}
    \hline
         Dataset &Sample size  &Comment\\
         \hline
          ZSL-and-Harvested-DSD-Tweets &2491 &Second harvest, sample size is same as Harvested-DSD-Tweets because harvesting is done on the same data\\
          Only-ZSL-Pred-on-Harvested-DSD-Tweets &2248 &Sample size less than the above because we are not using samples with no labels predicted\\
         \hline
    \end{tabular}
    \caption{\label{tab:harvested_datasets_2}Datasets in step 3}
   
\end{table}

\begin{table}%[!h]
    % \small
    \centering
    \begin{tabular}{|p{5cm}|c|c|c|c|}
    \hline
        Comment                  &Precision  &Recall &F1-score   &Support\\
        \hline
        Anhedonia                 &0.00      &0.00      &0.00         &5\\
        Low mood                  &0.56      &0.85      &0.68        &26\\
        Change in sleep pattern   &0.72      &0.87      &0.79        &15\\
        Fatigue                   &0.00      &0.00      &0.00         &6\\
        Weight change             &0.00      &0.00      &0.00         &4\\
        Feelings of worthlessness &0.33      &0.55      &0.42        &38\\
        Indecisiveness            &0.00      &0.00      &0.00        &11\\
        Agitation                 &1.00      &0.11      &0.19        &66\\
        Retardation               &0.00      &0.00      &0.00        &12\\
        Suicidal thoughts         &0.82      &0.64      &0.72        &22\\
        \hline
                    %Micro avg     &0.62      &0.51      &0.56       &205\\
                    Macro avg     &0.34      &0.30      &0.28       &205\\
                    Weighted avg  &0.60      &0.38      &0.36       &205\\
                    %Samples avg   &0.60      &0.57      &0.57       &205\\
        \hline
    \end{tabular}
    \caption{\label{tab:only_zsl_on_harvested_dsd_model_accuracy} DSD-Clinician-1 model accuracy in step 3a}
\end{table}

\begin{table}%[!h]
    % \small
    \centering
    \begin{tabular}{|p{5cm}|c|c|c|c|}
    \hline
        Comment                  &Precision  &Recall &F1-score   &Support\\
        \hline
        Anhedonia                 &0.00      &0.00      &0.00         &5\\
        Low mood                  &0.71      &0.92      &0.80        &26\\
        Change in sleep pattern   &0.68      &0.87      &0.76        &15\\
        Fatigue                   &0.00      &0.00      &0.00         &6\\
        Weight change             &0.00      &0.00      &0.00         &4\\
        Feelings of worthlessness &0.34      &0.82      &0.48        &38\\
        Indecisiveness            &0.00      &0.00      &0.00        &11\\
        Agitation                 &0.65      &0.82      &0.72        &66\\
        Retardation               &0.00      &0.00      &0.00        &12\\
        Suicidal thoughts         &0.76      &0.73      &0.74        &22\\
        \hline
                    %Micro avg     &0.62      &0.51      &0.56       &205\\
                    Macro avg     &0.31      &0.42      &0.35       &205\\
                    Weighted avg  &0.49      &0.67      &0.56       &205\\
                    %Samples avg   &0.60      &0.57      &0.57       &205\\
        \hline
    \end{tabular}
    \caption{\label{tab:zsl_and_harvested_dsd_model_accuracy} DSD-Clinician-1 model accuracy in step 3b}
\end{table}

\subsubsection{Step 4: Creating a second DSD Model:}\label{subsubsec:ssl-second-dsd-clinician}
From the previous experiments, we now create our second DSD model by retraining it with DSD-Clinician-Tweets-Original-Train and ZSL-and-Harvested-DSD-Tweets. This results in our second DSD model (Table \ref{tab:dsd_clinician_2_model}).

\begin{table}%[!h]
    \centering
    \begin{tabular}{|c|p{3cm}|p{4cm}|p{1.5cm}|}
    \hline
         Model &Train dataset &Sample size  &Comment\\
         \hline
         DSD-Clinician-2 &DSD-Clinician-Tweets-Original-Train + ZSL-and-Harvested-DSD-Tweets  &$(377+2491)=2868$  &DSD model at SSL iteration 2\\
         \hline
    \end{tabular}
    \caption{\label{tab:dsd_clinician_2_model}Model details in step 4}
\end{table}

\subsubsection{Step 5: Creating final DSD model} \label{subsubsec:ssl-final-dsd-clinician}
In this final step, we do the following:
\begin{enumerate}
    \item We create a combined dataset from D2S and DSD-Clinician-ED-Tweets and we call this combined dataset \textbf{DSD-Less-Confident-Tweets} dataset (Arrows 15, 16, 17, 20 in Figure \ref{fig:brief-ssl}). D2S tweets are used here because the dataset was annotated externally with a weak clinical annotation guideline. We use our model to further filter this dataset.
   
    \item We use DSD-Clinician-2 model and ZSL to harvest depression symptoms tweets from DSD-Less-Confident-Tweets, we call this dataset \textbf{ZSL-and-Harvested-DSD-from-Less-Confident-Tweets}. Finally with this harvested data and the datasets used to train DSD-Clinician-2 model, we create our final dataset called Final-DSD-Clinician-Tweets and by training with it, we learn our final DSD model called, Final-DSD-Clinician. We also retrain our DPD-Human model to create Final-DPD-Human model.  Datasets, models and the relevant statistics are reported in Tables \ref{tab:final_clinician_dataset}, \ref{tab:final_dsd_clinician_models} and \ref{tab:final_dsd_clinician_model_accuracy}.

\begin{figure}%[!h]
    \centering
    \includegraphics[width=1 \textwidth]{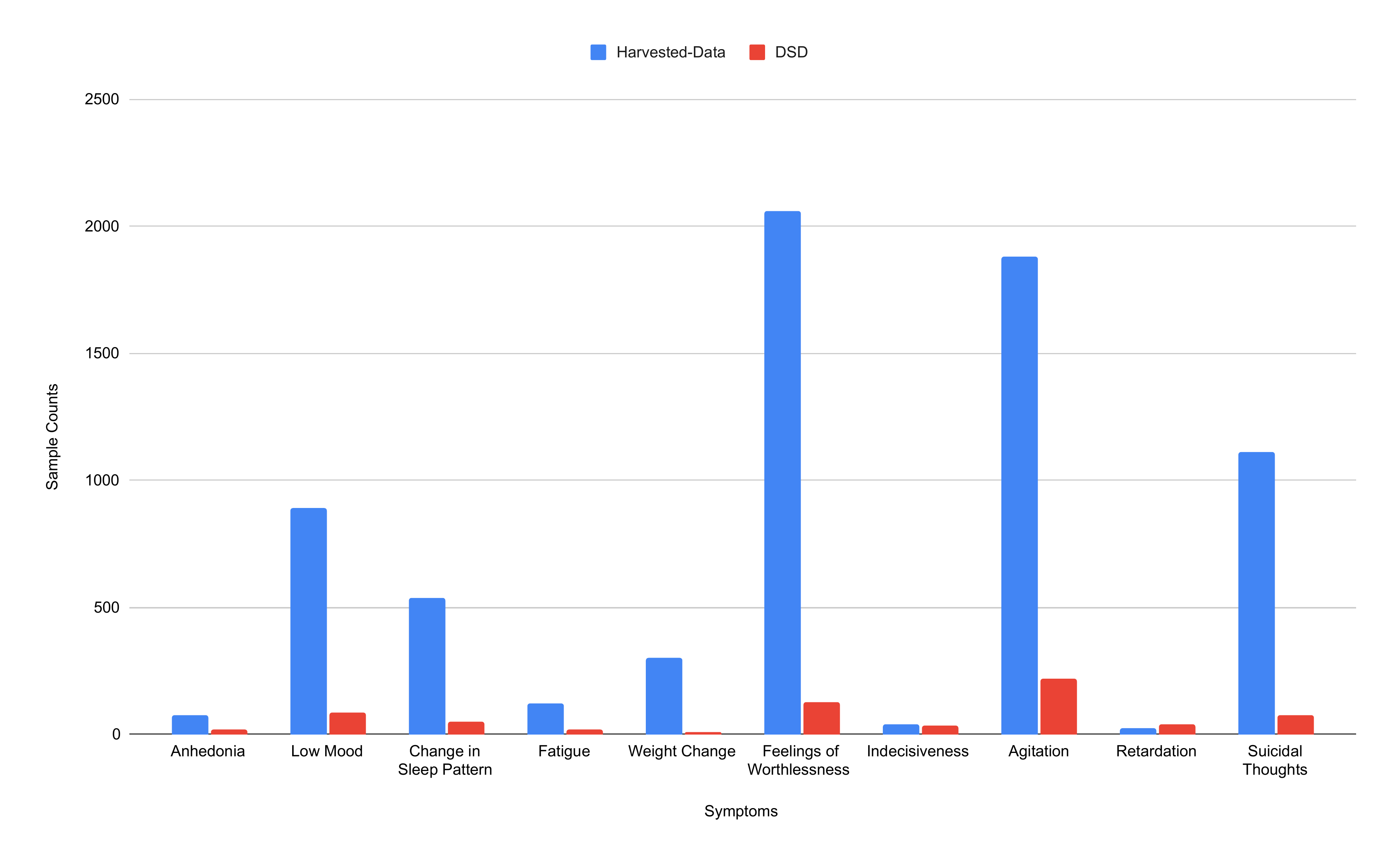}
    \caption{\label{fig:harvested-dataset-distr}Sample distribution in harvested dataset vs original clinician annotated dataset}
\end{figure}

We report the symptoms distribution for our DSD-Clinician-Tweets-Original-Train dataset earlier, and herereport depression symptoms distribution in our SSL model harvested datasets (ZSL-and-Harvested-DSD-Tweets + ZSL-and-Harvested-DSD-from-Less-Confident-Tweets) only. We see that sample size for all the labels generally increased and reflect almost the same distribution as our DSD-Clinician-Tweets-Original-Train dataset. Interestingly, data harvesting increase the sample size of ``Feelings of Worthlessness'' and ``Suicidal thoughts'' while still maintaining the distribution of our original clinician annotated dataset (DSD-Clinician-Tweets-Original-Train) (Figure \ref{fig:harvested-dataset-distr}).

%This captures the fact that, the general distribution from both D2S and DSD-Clinician-Tweets-Original-Train. 

We also report the top-10 bi-grams for each of the symptoms for our Final-DSD-Clinician-Tweets dataset in Table \ref{tab:bigrams-final-dsd}. We see that top bi-grams convey the concepts of each symptoms.
    
    \begin{table}%[!h]
    \centering
        \begin{tabular}{|p{3cm}|p{6cm}|p{2cm}|}
        \hline
             Dataset &Constituent datasets &Sample size \\% &Comment\\
             \hline
              Final-DSD-Clinician-Tweets &DSD-Clinician-Tweets-Original-Train + ZSL-and-Harvested-DSD-Tweets + ZSL-and-Harvested-DSD-from-Less-Confident-Tweets & $(377+2491+1699)=4567$ \\
              
              Final-DPD-Human-Tweets 
              & Final-DSD-Clinician-Tweets which are not in DPD-Human-1 testset + DPD-Human-1 trainset which are not in Final-DSD-Clinician-Tweets + Equal number of NoED tweets from DSD-Harvest-Candidates  &$(2743+1997) \times 2 = 9480$  \\ 
              %& Our DTR consists of 6077 samples, so for $(6077-4567)=1510$ samples neither ZSL nor any version of DSD models have any predictions, see arrow heads 18 and 22. \\
             \hline
        \end{tabular}
        \caption{\label{tab:final_clinician_dataset}Datasets in step 5}
    \end{table}

    \begin{table}%[!h]
        \centering
        \begin{tabular}{|c|p{5cm}|p{2.5cm}|}
        \hline
             Model &Train dataset &Comment\\
             \hline
               Final-DSD-Clinician &Final-DSD-Clinician-Tweets &DSD model at SSL Step 5\\
               Final-DPD-Human &Final-DPD-Human-Tweets &DPD model at SSL step 5\\
             \hline
        \end{tabular}
        \caption{\label{tab:final_dsd_clinician_models}Model details in step 5}
    \end{table}

    % this table will hold accuracy of step 3(a) in the Steps_and_Results file.
    \begin{table}%[!h]
        % \small
        \centering
        \begin{tabular}{|p{5cm}|c|c|c|c|}
        \hline
            Comment                  &Precision  &Recall &F1-score   &Support\\
            \hline
            Anhedonia                 &0.00      &0.00      &0.00         &5\\
            Low mood                  &0.57      &0.96      &0.71        &26\\
            Change in sleep pattern   &0.68      &0.87      &0.76        &15\\
            Fatigue                   &1.00      &0.17      &0.29         &6\\
            Weight change             &1.00      &0.75      &0.86         &4\\
            Feelings of worthlessness &0.35      &0.76      &0.48        &38\\
            Indecisiveness            &0.00      &0.00      &0.00        &11\\
            Agitation                 &0.62      &0.77      &0.69        &66\\
            Retardation               &0.00      &0.00      &0.00        &12\\
            Suicidal thoughts         &0.64      &0.82      &0.72        &22\\
            \hline
                        %Micro avg     &0.62      &0.51      &0.56       &205\\
                        Macro avg     &0.49      &0.51      &0.45       &205\\
                        Weighted avg  &0.51      &0.68      &0.56       &205\\
                        %Samples avg   &0.60      &0.57      &0.57       &205\\
            \hline
        \end{tabular}
        \caption{\label{tab:final_dsd_clinician_model_accuracy}Final-DSD-Clinician model accuracy in step 5}
    \end{table}

    \begin{table}%[!h]
        % \small
        \centering
        \begin{tabular}{|c|c|c|c|}
        \hline
                Precision  &Recall &F1-score   &Support\\
        \hline
                0.83       &0.97   &0.89       &227 \\
        \hline
        \end{tabular}
        \caption{\label{tab:final_dpd_human_model_accuracy}Final-DPD-Human model accuracy in step 5}
    \end{table}

    \begin{table}%[!ht]
    \centering
    \begin{tabular}{|l|p{8cm}|}
    \hline
        \textbf{Depression-Symptoms} & \textbf{Bi-grams} \\ \hline
        Anhedonia & want go, dont care, go work, motivation anything, want die, want live, go away, im done, tired bored, getting bed \\ \hline
        Low Mood & feel like, want cry, depression anxiety, feeling like, mental illness, want die, like shit, want someone, feel alone, feels like \\ \hline
        Change in Sleep Pattern & want sleep, go sleep, im tired, hours sleep, fall asleep, cant sleep, need sleep, back sleep, could sleep, going sleep \\ \hline
        Fatigue & im tired, f*cking tired, physically mentally, tired everything, tired tired, feel tired, im f*cking, need break, tired yall, sad tired \\ \hline
        Weight Change & eating disorder, fat fat, stop eating, feel like, keep eating, im gonna, lose weight, eating disorders, fat body, wish could \\ \hline
        Feelings of Worthlessness & feel like, like shit, feeling like, fat fat, wish could, f*cking hate, good enough, ibs hate, hate ibs, makes feel \\ \hline
        Indecisiveness & cant even, even know, says better, thoughts brain, seems like, feel like, better dead, assistant remember, remember things, time like \\ \hline
        Agitation & feel like, mental illness, f*ck f*ck, depression anxiety, f*ck life, f*cking hate, fat fat, panic attacks, every time, hate body \\ \hline
        Retardation & feel like, lay bed, ever get, committed bettering, sleepy kind, im tired, one moods, talking going, well mind, motherf*ckers prove \\ \hline
        Suicidal thoughts & want die, feel like, wanna die, want kill, want cut, f*cking die, better dead, self harm, hope die, want f*cking \\ \hline
        \end{tabular}
        \caption{\label{tab:bigrams-final-dsd}Top-10 bi-grams for each symptoms for Final-DSD-Clinician-Tweets dataset}
    \end{table}
    
    \begin{table}%[!h]
        % \small
        \centering
        \begin{tabular}{|p{5cm}|c|c|c|c|}
        \hline
            Comment                  &Precision  &Recall &F1-score   &Support\\
            \hline
            Anhedonia                 &0.03      &0.80      &0.06         &5\\
            Low mood                  &0.59      &0.92      &0.72        &26\\
            Change in sleep pattern   &0.71      &1.00      &0.83        &15\\
            Fatigue                   &0.04      &0.83      &0.08         &6\\
            Weight change             &1.00      &0.50      &0.67         &4\\
            Feelings of worthlessness &0.34      &0.79      &0.47        &38\\
            Indecisiveness            &0.09      &1.00      &0.16        &11\\
            Agitation                 &0.61      &0.76      &0.68        &66\\
            Retardation               &0.07      &0.75      &0.12        &12\\
            Suicidal thoughts         &0.72      &0.82      &0.77        &22\\
            \hline
                        %Micro avg     &0.62      &0.51      &0.56       &205\\
                        Macro avg     &0.42      &0.82      &0.45       &205\\
                        Weighted avg  &0.49      &0.82      &0.57       &205\\
                        %Samples avg   &0.60      &0.57      &0.57       &205\\
            \hline
        \end{tabular}
        \caption{\label{tab:final_apriori_dsd_model_accuracy}Final-DSD-Clinician model with applied label association rules accuracy in step 6}
    \end{table}

    \begin{table}%[!h]
        % \small
        \centering
        \begin{tabular}{|p{5cm}|c|c|c|c|}
        \hline
            Comment                  &Precision  &Recall &F1-score   &Support\\
            \hline
            Anhedonia                 &0.00      &0.00      &0.00           &5\\
            Low mood                  &0.52      &0.96      &0.68           &26\\
            Change in sleep pattern   &0.71      &1.00      &0.83           &15\\
            Fatigue                   &1.00      &0.17      &0.29           &6\\
            Weight change             &1.00      &0.75      &0.8            &4\\
            Feelings of worthlessness &0.32      &0.82      &0.46           &38\\
            Indecisiveness            &0.00      &0.00      &0.00           &11\\
            Agitation                 &0.64      &0.76      &0.69           &66\\
            Retardation               &0.00      &0.00      &0.00           &12\\
            Suicidal thoughts         &0.60      &0.82      &0.69           &22\\
            \hline
                        Macro avg     &0.48      &0.53      &0.45           &205\\
                        Weighted avg  &0.50      &0.70      &0.56           &205\\
            \hline
        \end{tabular}
        \caption{\label{tab:unlabelled_dataset_dsd_model_accuracy}DSD-Clinician model trained on IJCAI-2017-Unlabelled and all the harvested dataset}
    \end{table}

\end{enumerate}
    
\subsubsection{Step 6: Combating low accuracy for less populated labels} \label{subsubsec:ssl-combat-low-acc}
\begin{figure}%[!h]
    \centering
    \includegraphics[width=1 \textwidth]{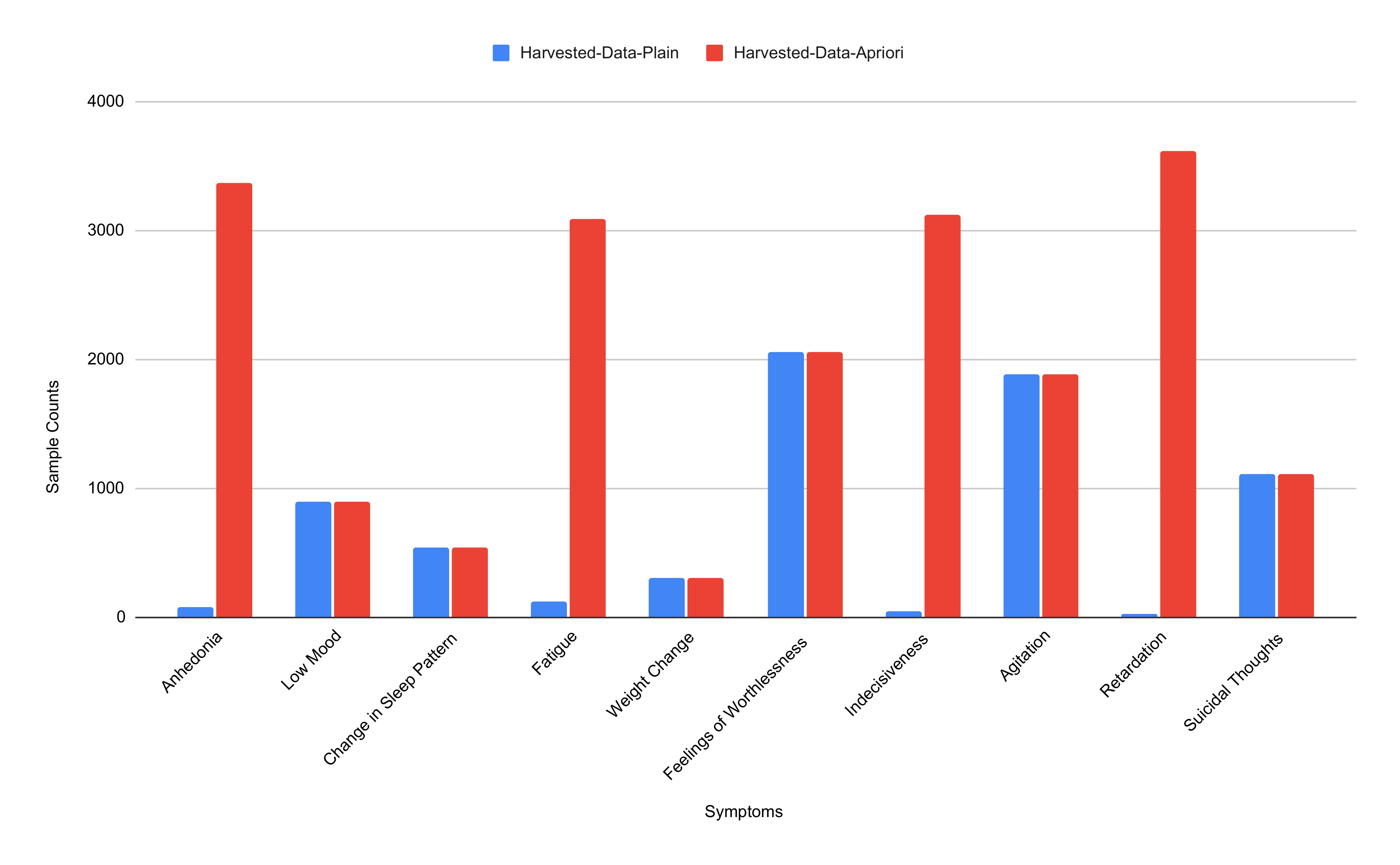}
    \caption{Sample distribution in Apriori harvested dataset vs plain harvested dataset\label{fig:aprior-vs-plain-distr}}
\end{figure}

Here we attempt to combat the low accuracy for the labels which have very small sample size. In these cases, we analyze the co-occurrence of those labels with other labels through an associative rule mining (Apriori) algorithm \cite{agrawal1994fast}. Our idea is to use significant co-occurring labels and artificially predict one label if the other occurs. For that, we analyze a small human annotated train dataset (DSD-Clinician-Tweets-Original-Train). However, since the support and confidence for association rules are not significant due to the small sample size, we consider all the ``strong'' rules with non-zero support and confidence score for those labels. The rules we consider have the form: (strong-label $\rightarrow$ weak-label), where the weak label (such as, Anhedonia, Fatigue, Indecisiveness and Retardation) means, the labels for which our model achieves either 0 F1 score or very low recall). These are the candidate labels for which we would like to have increased accuracy. On the other hand strong labels are those for which we have at least a good recall. By emphasizing high recall, we intend to not miss a depression symptom from being detected by our model. All the extracted strong rules are provided in Appendix \ref{appen:apriori-rules}. When we compare the sample distribution for Apriori based harvested data and plain harvested data, we see for the least populated class we have more samples (Figure \ref{fig:aprior-vs-plain-distr}). This makes the classification task more sensitive towards the weak labels. However, with this method, we do not achieve better Macro-F1 score compared to our Final-DSD-Clinician model (Table \ref{tab:final_apriori_dsd_model_accuracy}).

\subsubsection{Stopping criteria for SSL:}\label{subsubsec:ssl-stop-criteria}
The following two observations lead us to stop the SSL:
\begin{enumerate}
    \item Our DTR consists of total 6077 samples and we have finally harvested 4567 samples, so for $(6077-4567)=1510$ samples neither ZSL nor any version of DSD models have any predictions. We exhausted all our depression candidate tweets from all sources we have, therefore, we do not have any more depression symptoms candidate tweets for moving on with SSL.
    \item We have another very noisy dataset, called IJCAI-2017-Unlabelled \cite{shen2017depression}, where we have tweets from possible depressed users, i.e., their self-disclosure contains the stem ``depress'' but it is not verified whether they are genuine self-disclosures of depression. Using our Final-DSD-Clinician model we harvest $\approx 22K$ depression symptoms tweets from $\approx 0.4M$ depression candidate tweets identified by Final-DPD-Human model from that dataset. We then 
    %repeat the steps we followed at final steps described above and 
    retrain the Final-DSD-Clinician model on all the samples previously we harvested combined with the newly harvested $\approx 22K$ tweets, which results in a total of $\approx 26k$ tweets ($\approx 6$ times larger than the samples DSD-Final-model was trained on). However, we did not see any significant accuracy increase, so we did not proceed (Table \ref{tab:unlabelled_dataset_dsd_model_accuracy}). 
\end{enumerate}

\section{Results Analysis} \label{sec:ssl-results}
Here we analyse the efficacy of our semi-supervised learning frameworks on three dimensions, as follows:

\subsection{Dataset size increase} \label{subsec:ssl-dataset-increase} Through the data harvesting process, we are able to increase our initial clinician annotated 377 samples to 4567 samples, which is 12 times bigger than our initial dataset. In addition, we have access to a external organization collected dataset, for which we could access around $\approx 1800$ samples. Our final dataset is more than double the size of that dataset.

\subsection{Accuracy improvement} \label{subsec:ssl-acc-improve} Our Final-DSD-Clinician model has Macro-F1 score of 45\% which is 14\% more than that of our initial model and Weighted-F1 score increased by 5\% from 51\% to 56\%.  The substantial gain in Macro-F1 score indicates the efficacy of our data harvesting in increasing F1 scores for all the labels.  We also find that the combination of DSD-Clinician-1 and ZSL models in step 3a helps achieve more accuracy than individually; specifically, using only ZSL harvested data for training is not very ideal.  Micro-F1 has slow growth and does not increase after Step 3b.  We also find that the combined harvesting process on D2S sampled helped us achieve further accuracy in a few classes for which D2S had more samples, such as ``Fatigue,'' ``Weight Change'' and ``Suicidal Thoughts.''

%Substantial gain in Macro-F1 score indicates the efficacy of our data harvesting for increasing F1 scores for all the labels. We also find, combination of DSD-Clinician-1 and ZSL models in step: 3a, helps achieve more accuracy than individually, specifically only ZSL harvested data for training is not very ideal.  Micro-F1 has slow growth; after step: 3b it does not increase. We also find combined harvesting process on D2S samples help us achieve further accuracy in few classes such as for which D2S has more samples, such as, ``Fatigue'', ``Weight Change'' and ``Suicidal thoughts''.
%This means, harvesting from DSD-Less-Confident-Tweets does not help make labels with majority samples even better, but has impact on labels with small samples. 

\subsection{Linguistic components distribution} \label{subsec:ssl-ling-comp-distr} In Table \ref{tab:bigrams-final-dsd}, we see that our harvested dataset contains important clues of depression symptoms. Interestingly, there are some bi-grams, such as, ``feel like'' occurs in most of the labels; this signifies the frequent usage of that bi-gram in various language based expressions of depression symptoms. This also shows a pattern of how people describe their depression.

\subsection{Sample distribution} \label{subsec:ssl-samp-distr} Compared with the original clinician annotated dataset distribution (Table \ref{fig:aprior-vs-plain-distr}), we see similar trends in our harvested dataset, i.e., in Final-DSD-Clinician-Tweets. However, instead of ``Agitation'' we have some more samples on ``Feeling of Worthlessness,'' although those are not surpassed by ``Suicidal thoughts'' as in D2S dataset. Moreover, ``Suicidal thoughts'' samples have also strong presence which is the result of integrating D2S dataset in our harvesting process. Since the majority of our samples are coming from self-disclosed users tweets, and we apply our DSD model learned on that dataset to the D2S dataset to harvest tweets, our final harvested dataset reflects mainly the distribution of symptoms from the self-disclosed depressed users. However, D2S has some impact which results in more samples in the most populated labels of the final harvested dataset. 

\subsection{Data harvesting in the wild} \label{subsec:ssl-harvest-wild} We use our final model on a bigger set of very loosely related data, but we do not see any increase of accuracy, which suggests that harvesting from irrelevant data is of no use.

\section{Limitations} \label{sec:ssl-limit}
\begin{enumerate}
    \item Our overall dataset size is still small, i.e. for some labels we have very small amount of data both for training and testing.
    % \item We do not attempt to artificially increase sample amount for the small populated labels.
    % \item To start with, we use a small expert (i.e. clinician) annotated dataset because human annotation is expensive.
    % \item We haven't explored stratified sample generation using state-of-the-art generative models.
    \item In the iterative harvesting process we do not employ continuous human annotation or human-in-the-loop strategy, since this process requires several such cycles and involving experts in such framework is also very expensive.
\end{enumerate}

\section{Conclusion}
\label{sec:ssl-conclusion}
We have described a Semi-supervised Learning (SSL) framework, more specifically semi-supervised co-training for gathering depression symptoms data in-situ from self-disclosed users Twitter timeline. We articulate each step of our data harvesting process and model re-training process. We also discuss our integration of Zero-Shot learning models in this process and their contribution.  We show that each of these steps provides moderate to significant accuracy gains. We discuss the effect of harvesting from the samples of an externally curated dataset, and we also try harvesting samples in the wild, i.e., a large noisy dataset with our Final-DSD-Clinician model. In the former case we find good improvement in Macro-F1 score. In the latter, we do not see any improvements indicating that there is room for further progress to improve accuracy in those samples. Finally, we discuss the effect of our SSL process for curating small but distributionally relevant samples through both sample distribution and bi-gram distribution for all the labels.

%% author's contribution, ethical information, conflict of interest, acknowledgment, and funding 

\section{Data Availability Statement}
The datasets generated and/or analysed during the current study are not yet publicly available due to the privacy and ethical implications regarding the identity of Twitter users. However, we are working actively to release the dataset soon either in encoded form, i.e. through a sentence embedding representation or paraphrased form.

\section{Authors Contribution}
N.F. developed the original research idea, designed and conducted experiments, annotated samples, wrote and reviewed the manuscript.
R.G. Reviewed the manuscript and managed funding.
S.S. Helped in creating annotation guideline, annotated samples and reviewed the manuscript.
O.Z. Reviewed the manuscript.

\section{Ethical Information} We obtained ethics approval from University of Alberta's research ethics office for ``Depression Detection from Social Media Language Usage'' (Pro00099074), ``Depression Dataset Collection'' (Pro00082738) and ``Social Media Data Annotation by Human'' (Pro00091801). 

\section{Data Availability Statement}
The datasets generated and/or analysed during the current study are not yet publicly available due to the privacy and ethical implications regarding the identity of Twitter users. However, we are working actively to release the dataset soon either in encoded form, i.e. through a sentence embedding representation or paraphrased form.

% \section{Conflict of Interest} There is no known conflict of interest.

\section{Acknowledgment} We are grateful to the annotators, Katrina Regan-Ingram, MA, Communications and Technology and Jasmine M. Noble, PhD,  Psychiatry, University of Alberta for allocating their time for data annotation. 

\section{Funding} This research is supported by a grant from Alberta Machine Intelligence Institute (AMII). Grant no is: RES0042153.

\begin{appendices}

\section{Apriori rules}
\label{appen:apriori-rules}
Here we provide the strong rules mined from DSD-Clinician-Tweets-Original-Train (Table \ref{tab:apriori-rules})

\begin{table}[!h]
    % \small
    \centering
    \begin{tabular}{|p{5cm}|}
    \hline
    Rules(Strong-Label $\rightarrow$ Weak-Label)\\
    \hline
    1 $\rightarrow$ 2\\
    1 $\rightarrow$ 6\\
    \hline
    4 $\rightarrow$ 3\\
    4 $\rightarrow$ 8\\
    4 $\rightarrow$ 10\\
    \hline
    7 $\rightarrow$ 6\\
    7 $\rightarrow$ 8\\
    \hline
    9 $\rightarrow$ 6\\
    9 $\rightarrow$ 8\\
    9 $\rightarrow$ 10\\
    \hline
    \end{tabular}
    \caption{\label{tab:apriori-rules} Strong Rules; indices for each labels are from Section \ref{appen:annots-guideline}}
\end{table}

\section{Mental-BERT Training Configuration for DPD and DSD}
Here we report the training configuration for Mental-BERT based DPD and DSD (Table \ref{tab:train-params})
\label{appen:train-params}

\begin{table}[!h]
    % \small
    \centering
    \begin{tabular}{|c|c|c|}
    \hline
            Hyperparameters &DPD                    &DSD\\
    \hline
            \#Epochs            &20                 &10\\
            \#Batch             &32                 &Same\\
            MAX sequence length &30                 &Same\\
            Learning Rate       &$2\times10^{-5}$   &Same\\
            \#GPUs              &1                  &Same\\
            Loss function       &Binary Cross Entropy(BCE) Loss            &Same\\
    \hline
    \end{tabular}
    \caption{\label{tab:train-params} DPD and DSD Model Training Parameters}
\end{table}

For DSD we use BCE Loss on the output of last layer of our Mental-BERT model which is based on sigmoid functions for each nodes corresponding to each depression symptoms labels. For DPD, we use BCE loss on the softmaxed output for each binary labels i.e. depression vs control. We do not freeze any layers in our fine-tuning process because it turned out to be detrimental to the model accuracy. 

\section{Annotation Guideline}
\label{appen:annots-guideline}
\subsection{Social Media Data Annotation by Human}

For this annotation task, an annotator has to label or classify a social media post (i.e. a tweet) in one or more of the following depression symptom categories which suit best for that social media post through a web tool:

\begin{enumerate}
    \item  Inability to feel pleasure or Anhedonia
    \item  Low mood 
    \item  Change in sleep pattern
    \item  Fatigue or loss of energy
    \item  Weight change or change in appetite 
    \item  Feelings of worthlessness or excessive inappropriate guilt
    \item  Diminished ability to think or concentrate or indecisiveness 
    \item  Psychomotor Agitation or Inner Tension 
    \item  Psychomotor Retardation 
    \item  Suicidal Thoughts or Self-Harm
    \item  Evidence of Clinical Depression
    \item  No evidence of Clinical Depression
    \item  Gibberish
\end{enumerate}

Detailed description of these categories with examples are as follows:

The following sections need to be very carefully read to better understand what each category means. We divide the description under each category into three parts: ``Lead'', ``Elaboration'', and ``Example''.  ``Lead'' contains the summary or gist of the symptomatology.  ``Elaboration'' provide a broader description of the symptomatology accompanied by a few relevant ``Examples''. These sections have been developed with careful considerations of criteria defined in the DSM-5 and MADRS, BDI, CES-D and PHQ-9 depression rating scales.
%into 3 parts, Lead, Elaboration and Example. Lead contains the summary or gist of the symptomatology. Elaboration as the name suggests, provides broader picture of the symptomatology accompanied by few relevant Examples under “Example”. 

\subsection{Depression Symptoms Labels}
    \begin{enumerate}
        \item  \textbf{Inability to feel pleasure or anhedonia}
        \begin{enumerate}
            \item \textbf{Lead:} Subjective experience of reduced interest in the surroundings or activities, that normally give pleasure.
            \item \textbf{Elaboration:} Dissatisfied and bored about everything. Not enjoying things as one would used to. Not enjoying life. Lost Interest in other people. Lost interest in sex. Can't cry anymore even though one wants to.
            \item \textbf{Example:} 
                    \begin{enumerate}
                        \item I feel numb.
                        \item I am dead inside. 
                        \item I don’t give a damn to anything anymore.
                    \end{enumerate}
        \end{enumerate}
    %\end{enumerate}
    
    %\begin{enumerate}
        \item  \textbf{Diminished ability to think or concentrate or indecisiveness}
        \begin{enumerate}
            \item \textbf{Lead:}  Difficulties in collecting one's thoughts mounting to incapacitating lack of concentration.
            \item \textbf{Elaboration:}  Can't make decisions at all anymore. Trouble keeping one's mind on what one was doing. Trouble concentrating on things.
            \item \textbf{Example:} 
                    \begin{enumerate}
                        \item I can't make up my mind these days.
                    \end{enumerate}
        \end{enumerate}
    %\end{enumerate}
    
    %\begin{enumerate}
        \item \textbf{Change in sleep pattern}
        \begin{enumerate}
            \item \textbf{Lead:} Reduced duration or depth of sleep, or increased duration of sleep compared to one's normal pattern when well.  
            \item \textbf{Elaboration:} Trouble Falling or Staying Asleep. Waking up earlier and cannot go back to sleep. Sleep was restless (wake up not feeling rested). Sleeping too much.
            \item \textbf{Example:} 
                    \begin{enumerate}
                        \item It's 3 am, and I am still awake.
                        \item I sleep all day!
                    \end{enumerate}
        \end{enumerate}
    %\end{enumerate}
    
    %\begin{enumerate}
        \item \textbf{Fatigue or loss of energy}
        \begin{enumerate}
            \item \textbf{Lead:} Any physical manifestation of tiredness.  
            \item \textbf{Elaboration:} Elaboration: Feeling tired. Insufficient energy for tasks. Feeling too tired to do anything.
            \item \textbf{Example:} 
                    \begin{enumerate}
                        \item I feel tired all day.
                        \item I feel sleepy all day.
                        \item I get exhausted very easily.
                    \end{enumerate}
        \end{enumerate}
    %\end{enumerate}
    
    %\begin{enumerate}
        \item \textbf{Feelings of worthlessness or excessive inappropriate guilt}
        \begin{enumerate}
            \item \textbf{Lead:} Representing thoughts of guilt, inferiority, self-reproach, sinfulness, and self-depreciation.  
            \item \textbf{Elaboration:} Feeling like a complete failure, Feeling guilty, Feeling of being punished. Self-hate. Disgusted and Disappointed on oneself. Self blaming for everything bad happens. Believe that one looks ugly or unattractive. Having crying spells. Feeling lonely. People seems unfriendly. Felt like all other people dislike oneself.  
            \item \textbf{Example:}  
                    \begin{enumerate}
                        \item Leave me alone, I want to go somewhere where there is no one.
                        \item I am so alone ...
                        \item Everything bad happens, happens because of me.
                    \end{enumerate}
        \end{enumerate}
    %\end{enumerate}
    
    %\begin{enumerate}
        \item \textbf{Low mood}
        \begin{enumerate}
            \item \textbf{Lead:}  Despondency, Gloom, Despair, Depressed Mood, Low Spirits, Feeling of being beyond help without hope.
            \item \textbf{Elaboration:} Feeling down. Feeling sad. Discouraged about future. Hopelessness. Feeling like it's not possible to shake of the blues even with the help of family and friends.
            \item \textbf{Example:} 
                    \begin{enumerate}
                        \item Life will never get any better.
                        \item I don't know why but I feel so empty.
                        \item I am so lost.
                        \item There is no hope to get out of this bad situation.
                    \end{enumerate}
        \end{enumerate}
    %\end{enumerate}
    
    %\begin{enumerate}
        \item \textbf{Psychomotor agitation or inner tension}
        \begin{enumerate}
            \item \textbf{Lead:}  Ill defined discomfort, edginess, inner-turmoil, mental tension mounting to either panic, dread or anguish.
            \item \textbf{Elaboration:} Feeling irritated and annoyed all the time. Bothered by things that usually don’t bother. Feeling fearful. Feeling Restless. Feeling Mental Pain.
            \item \textbf{Example:} 
                    \begin{enumerate}
                        \item It's my life so I decide what to do next, mind your own business, don't bother! 
                        \item You have no idea how much pain you gave me!
                    \end{enumerate}
        \end{enumerate}
    %\end{enumerate}
    
    %\begin{enumerate}
        \item \textbf{Psychomotor retardation or lassitude}
        \begin{enumerate}
            \item \textbf{Lead:} Difficulty getting started or slowness initiating and performing everyday activities.
            \item \textbf{Elaboration:} Feeling everything one do requires effort. Could not get going. Talked less than usual. Have to push oneself to do anything. Everything is a struggle. Moving or talking slowly.
            \item \textbf{Example:} 
                    \begin{enumerate}
                        \item I don't feel like moving from the bed.
                    \end{enumerate}
        \end{enumerate}
    %\end{enumerate}
    
    %\begin{enumerate}
        \item \textbf{Suicidal thoughts or self-Harm}
        \begin{enumerate}
            \item \textbf{Lead:} Feeling of Life is not worth living, suicidal thoughts, preparation for suicide.
            \item \textbf{Elaboration:} Recurrent thoughts of death (not just fear of dying), recurrent suicidal ideation without specific plan, or suicide attempt, or a specific plan for suicide. Thoughts of self-harm. Suicidal ideation. Drug abuse. 
            \item \textbf{Example:} 
                    \begin{enumerate}
                        \item I want to leave for the good.
                        \item 0 days clean.
                    \end{enumerate}
        \end{enumerate}
    %\end{enumerate}
    
    %\begin{enumerate}
        \item \textbf{Weight change or change in appetite}
        \begin{enumerate}
            \item \textbf{Lead:} Loss or gain of appetite or weight than usual. 
            \item \textbf{Elaboration:} Increase in weight. Decrease in weight. Increase in appetite. Decrease in appetite. Do not feel like eating. Poor appetite. Loss of desire to food, forcing oneself to eat. Eating a lot but not feeling satiated. Eating even if one is full. Eating in large amount of food quickly and repeatedly. Difficulty in stop eating.
            \item \textbf{Example:} 
                    \begin{enumerate}
                        \item I think I am over eating these days! 
                        \item I don't feel like eating anything!
                    \end{enumerate}
        \end{enumerate}
    %\end{enumerate}
    
    %\begin{enumerate}
        \item \textbf{Evidence of clinical depression}
        \begin{enumerate}
            \item \textbf{Elaboration:} Any social media post  which do not necessarily fit into any of the above symptoms, however still carry signs of depression or representing many symptoms at a time, so it’s very hard to fit it in a few symptoms.
            \item \textbf{Example:} 
                    \begin{enumerate}
                        \item I feel like I am drowning … 
                    \end{enumerate}
        \end{enumerate}
    %\end{enumerate}
    
    %\begin{enumerate}
        \item \textbf{No evidence of clinical depression}
        \begin{enumerate}
            \item \textbf{Elaboration:} Political stance or personal opinion, inspirational statement or advice, unsubstantiated claim or fact.
            \item \textbf{Example:} 
                    \begin{enumerate}
                        \item People who eat dark chocolate are less likely to be depressed.
                    \end{enumerate}
        \end{enumerate}
    %\end{enumerate}
    
    %\begin{enumerate}
        \item \textbf{Gibberish}
        \begin{enumerate}
            \item \textbf{Elaboration:} If you are not sure what a social media post means i.e. if a social media post does not make sense or it’s gibberish, then annotate it as Gibberish.
        \end{enumerate}
    \end{enumerate}




\end{appendices}

\bibliography{refs}% common bib file


\end{document}